\documentclass[10pt,twocolumn,letterpaper]{article}

\usepackage{cvpr}
\usepackage{times}
\usepackage{epsfig}
\usepackage{graphicx}
\usepackage{amsmath}
\usepackage{amssymb}
\usepackage{comment}
\usepackage{caption}

\def \path{\bp C}

\def\vec#1{\mathchoice{\mbox{\boldmath$\displaystyle#1$}}
  {\mbox{\boldmath$\textstyle#1$}}
  {\mbox{\boldmath$\scriptstyle#1$}}
  {\mbox{\boldmath$\scriptscriptstyle#1$}}}
\newcommand{\geo}{{\mathrm{geo}}}
\newcommand{\alb}{{\mathrm{alb}}}
\newcommand{\id}{{\mathrm{id}}}
\newcommand{\expr}{{\mathrm{exp}}}
\newcommand{\col}{{\mathrm{col}}}
\newcommand{\lan}{{\mathrm{lan}}}
\newcommand{\reg}{{\mathrm{reg}}}
\newcommand*{\LOWRES}{}

\usepackage[breaklinks=true,bookmarks=false]{hyperref}

\cvprfinalcopy % *** Uncomment this line for the final submission

\def\cvprPaperID{592} % *** Enter the CVPR Paper ID here

\ifcvprfinal\fi

\usepackage{fancyhdr}
\fancypagestyle{plain}{%
	\fancyhf{}%
	\fancyfoot[C]{This is a preprint of the accepted version of the following CVPR2016 article: "Face2Face: Real-time Face Capture and Reenactment of RGB Videos".}
}

\begin{document}

%%%%%%%%% TITLE
\title{Face2Face: Real-time Face Capture and Reenactment of RGB Videos}

\author{%
	Justus Thies$^{1}$~~~~Michael Zollh\"ofer$^{2}$~~~~Marc Stamminger$^{1}$~~~~Christian Theobalt$^{2}$~~~~Matthias Nie{\ss}ner$^{3}$\vspace{0.1cm} \\ 
	$^{1}$University of Erlangen-Nuremberg~~~$^{2}$Max-Planck-Institute for Informatics~~~ $^{3}$Stanford University	
	\vspace{0.1cm}
}

\twocolumn[{%
	\renewcommand\twocolumn[1][]{#1}%
	\maketitle
	{\centering
		\vspace{-0.8cm}
		\includegraphics[width=\linewidth]{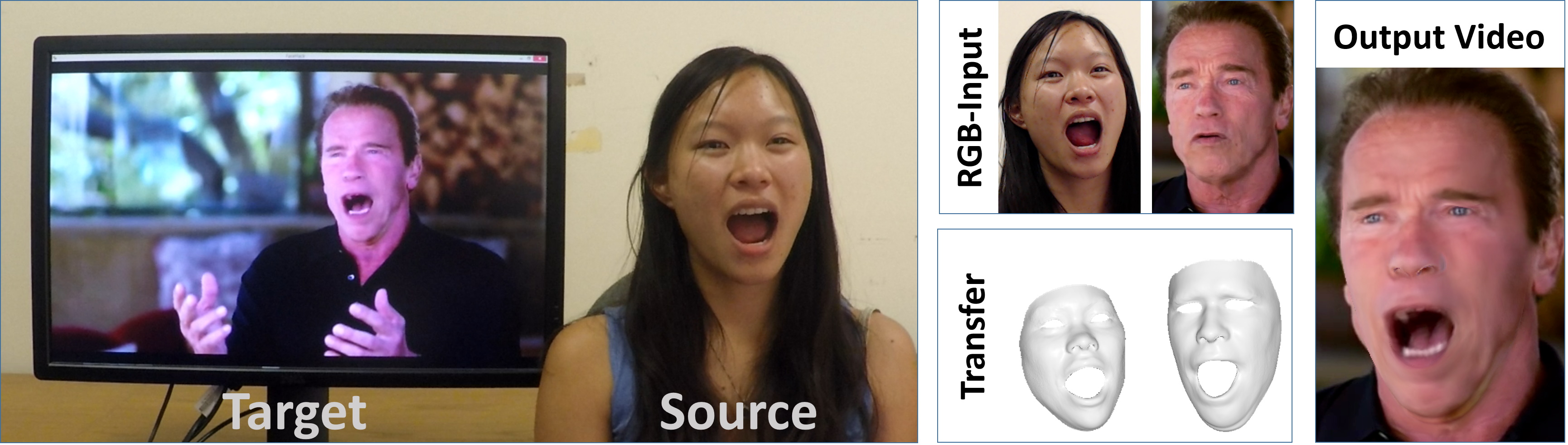}
	}
	Proposed online reenactment setup: a monocular target video sequence (e.g., from Youtube) is reenacted based on the expressions of a source actor who is recorded live with a commodity webcam.
	\label{fig:teaser}
	\vspace{0.5cm}
}]

\begin{abstract}
We present a novel approach for real-time facial reenactment of a monocular target video sequence (e.g., Youtube video).
The source sequence is also a monocular video stream, captured live with a commodity webcam.
Our goal is to animate the facial expressions of the target video by a source actor and re-render the manipulated output video in a photo-realistic fashion.  
To this end, we first address the under-constrained problem of facial identity recovery from monocular video by non-rigid model-based bundling.
At run time, we track facial expressions of both source and target video using a dense photometric consistency measure.
Reenactment is then achieved by fast and efficient deformation transfer between source and target.
The mouth interior that best matches the re-targeted expression is retrieved from the target sequence and warped to produce an accurate fit.
Finally, we convincingly re-render the synthesized target face on top of the corresponding video stream such that it seamlessly blends with the real-world illumination.
We demonstrate our method in a live setup, where Youtube videos are reenacted in real time.
\end{abstract}

\section{Introduction}

In recent years, real-time markerless facial performance capture based on commodity sensors has been demonstrated.
Impressive results have been achieved, both based on RGB \cite{Cao2013, Cao2015} as well as RGB-D data \cite{Weise2011,Chen:2013,Li2013,Bouaziz2013,hsieh2015unconstrained}.
These techniques have become increasingly popular for the animation of virtual CG avatars in video games and movies.
It is now feasible to run these face capture and tracking algorithms from home, which is the foundation for many VR and AR applications, such as teleconferencing.

In this paper, we employ a new dense markerless facial performance capture method based on monocular RGB data, similar to state-of-the-art methods.
However, instead of transferring facial expressions to virtual CG characters, our main contribution is monocular \emph{facial reenactment} in real-time.
In contrast to previous reenactment approaches that run offline \cite{Bregler1997,Dale2011,Garrido2014}, our goal is the \textit{online} transfer of facial expressions of a source actor captured by an RGB sensor to a target actor.
The target sequence can be any monocular video; e.g., legacy video footage downloaded from Youtube with a facial performance. 
We aim to modify the target video in a photo-realistic fashion, such that it is virtually impossible to notice the manipulations.
Faithful photo-realistic facial reenactment is the foundation for a variety of applications; for instance, in video conferencing, the video feed can be adapted to match the face motion of a translator, or face videos can be convincingly dubbed to a foreign language.

\begin{figure}
	\centering
	\ifdefined\LOWRES
	\includegraphics[width=0.9\linewidth]{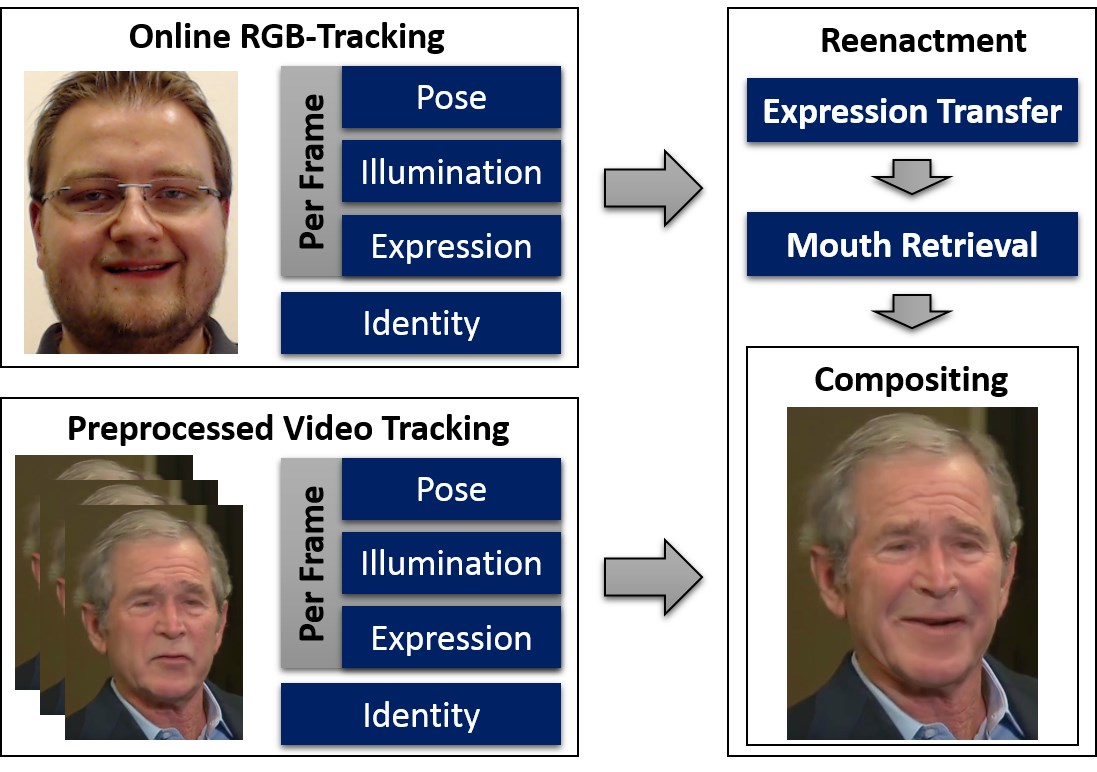}
	\else
	\includegraphics[width=0.9\linewidth]{images/pipeline2}
	\fi
	\caption{Method overview.}
	\label{fig::overview}
\end{figure}

In our method, we first reconstruct the shape identity of the target actor using a new global non-rigid model-based bundling approach based on a prerecorded training sequence.
As this preprocess is performed globally on a set of training frames, we can resolve geometric ambiguities common to monocular reconstruction.
At runtime, we track both the expressions of the source and target actor's video by a dense analysis-by-synthesis approach based on a statistical facial prior.
We demonstrate that our RGB tracking accuracy is on par with the state of the art, even with online tracking methods relying on depth data.
In order to transfer expressions from the source to the target actor in real-time, we propose a novel transfer functions that efficiently applies deformation transfer \cite{Sumner2004} directly in the used low-dimensional expression space.
For final image synthesis, we re-render the target's face with transferred expression coefficients and composite it with the target video's background under consideration of the estimated environment lighting.
Finally, we introduce a new image-based mouth synthesis approach that generates a realistic mouth interior by retrieving and warping best matching mouth shapes from the offline sample sequence.
It is important to note that we maintain the appearance of the target mouth shape;
in contrast, existing methods either copy the source mouth region onto the target \cite{Vlasic2005,Dale2011}
or a generic teeth proxy is rendered \cite{Garrido2015,thies2015realtime}, both of which leads to inconsistent results.
Fig.~\ref{fig::overview} shows an overview of our method.

%In the end, we are able to 
We demonstrate highly-convincing transfer of facial expressions from a source to a target video in real time.
We show results with a live setup where a source video stream, which is captured by a webcam, is used to manipulate a target Youtube video.
In addition, we compare against state-of-the-art reenactment methods, which we outperform both in terms of resulting video quality and runtime (we are the first real-time RGB reenactment method).
In summary, our key contributions are:

\begin{itemize}
	\addtolength{\itemsep}{-0.6\baselineskip}
	\item dense, global non-rigid model-based bundling,
	\item accurate tracking, appearance, and lighting estimation in unconstrained live RGB video,
	\item person-dependent expression transfer using subspace deformations,
	\item and a novel mouth synthesis approach.
\end{itemize}

\section{Related Work}
\label{sec:related_work}

\paragraph{Offline RGB Performance Capture}

Recent offline performance capture techniques approach the hard monocular reconstruction problem by fitting a blendshape \cite{Garrido2013} or a multi-linear face \cite{Shi2014} model to the input video sequence.
Even geometric fine-scale surface detail is extracted via inverse shading-based surface refinement.
Ichim et al. \cite{Ichim2015} build a personalized face rig from just monocular input.
They perform a structure-from-motion reconstruction of the static head from a specifically captured video, to which they 
fit an identity and expression model.
Person-specific expressions are learned from a training sequence.
Suwajanakorn et al. \cite{Suwajanakorn2014} learn an identity model from a collection of images and track the facial animation based on a model-to-image flow field.
Shi et al.~\cite{Shi2014} achieve impressive results based on global energy optimization of a set of selected keyframes.
Our model-based bundling formulation to recover actor identities is similar to their approach;
however, we use robust and dense global photometric alignment, which we enforce with an efficient data-parallel optimization strategy on the GPU.

\paragraph{Online RGB-D Performance Capture}

Weise et al.~\cite{weise09face} capture facial performances in real-time by fitting a parametric blendshape model to RGB-D data, but they require a professional, custom capture setup.
The first real-time facial performance capture system based on a commodity depth sensor has been demonstrated by Weise et al.~\cite{Weise2011}.
Follow up work \cite{Li2013,Bouaziz2013,Chen:2013,hsieh2015unconstrained} focused on corrective shapes \cite{Bouaziz2013}, dynamically adapting the blendshape basis \cite{Li2013}, non-rigid mesh deformation \cite{Chen:2013}, and robustness against occlusions \cite{hsieh2015unconstrained}.
These works achieve impressive results, but rely on depth data which is typically unavailable in most video footage. %do not deal with the problem of monocular RGB based tracking, which is a necessity for handling legacy video footage; e.g., from Youtube.

\paragraph{Online RGB Performance Capture}

While many sparse real-time face trackers exist, e.g., \cite{Saragih2011a}, real-time dense monocular tracking is the basis of realistic online facial reenactment. 
Cao et al.~\cite{Cao2013} propose a real-time regression-based approach to infer $3$D positions of facial landmarks which constrain a user-specific blendshape model.
Follow-up work~\cite{Cao2015} also regresses fine-scale face wrinkles.
These methods achieve impressive results, but are not directly applicable as a component in facial reenactment, since they do not facilitate dense, pixel-accurate tracking.

\paragraph{Offline Reenactment}
Vlasic et al.~\cite{Vlasic2005} perform facial reenactment by tracking a face template, which is re-rendered under different expression parameters on top of the target; the mouth interior is directly copied from the source video.
Dale et al.~\cite{Dale2011} achieve impressive results using a parametric model, but they target face replacement and compose the source face over the target.
Image-based offline mouth re-animation was shown in~\cite{Bregler1997}. Garrido et al.~\cite{Garrido2014} propose an automatic purely image-based approach to replace the entire face.
These approaches merely enable self-reenactment; i.e., when source and target are the same person; in contrast, we perform reenactment of a different target actor. 
Recent work presents virtual dubbing \cite{Garrido2015}, a problem similar to ours; however, the method runs at slow offline rates and relies on a generic teeth proxy for the mouth interior.
Kemelmacher et al.~\cite{Kemelmacher2011} generate face animations from large image collections, but the obtained results lack temporal coherence.
Li et al.~\cite{Li2012} retrieve frames from a database based on a similarity metric.
They use optical flow as appearance and velocity measure and search for the $k$-nearest neighbors based on time stamps and flow distance.
Saragih et al.~\cite{Saragih2011a} present a real-time avatar animation system from a single image.
Their approach is based on sparse landmark tracking, and the mouth of the source is copied to the target using texture warping.
Berthouzoz et al. \cite{Berthouzoz2012} find a flexible number of in-between frames for a video sequence using shortest path search on a graph that encodes frame similarity.
Kawai et al. \cite{Kawai2014} re-synthesize the inner mouth for a given frontal 2D animation using a tooth
and tongue image database; they are limited to frontal poses, and do not produce as realistic renderings as ours under general head motion.

\paragraph{Online Reenactment}

Recently, first online facial reenactment approaches based on RGB-(D) data have been proposed.
Kemelmacher-Shlizerman et al.~\cite{Kemelmacher-ShlizermanSSS10} enable image-based puppetry by querying similar images from a database.
They employ an appearance cost metric and consider rotation angular distance, which is similar to Kemelmacher et al.~\cite{Kemelmacher2011}.
While they achieve impressive results, the retrieved stream of faces is not temporally coherent.
Thies et al.~\cite{thies2015realtime} show the first online reenactment system; however, they rely on depth data and use a generic teeth proxy for the mouth region.
In this paper, we address both shortcomings: 1) our method is the first real-time RGB-only reenactment technique; 2) we synthesize the mouth regions exclusively from the target sequence (no need for a teeth proxy or direct source-to-target copy).

\section{Synthesis of Facial Imagery} 
\label{sec::synthesis}

We use a multi-linear PCA model based on \cite{Blanz1999,Alexander2009,Cao2014b}.
The first two dimensions represent facial identity -- i.e., geometric shape and skin reflectance -- and the third dimension controls the facial expression.
Hence, we parametrize a face as:
\begin{alignat}{2}
\mathcal{M}_\geo(\vec{\alpha}, \vec{\delta})  & \; = \;  \vec{a}_\id && + E_\id \cdot \vec{\alpha} + E_\expr \cdot \vec{\delta}~, \\
\mathcal{M}_\alb \left( \vec{\beta} \right)                & \; = \; \vec{a}_\alb && + E_\alb \cdot \vec{\beta}\enspace{.}
\end{alignat}
This prior assumes a multivariate normal probability distribution of shape and reflectance around the average shape $\vec{a}_\id \in \mathbb{R}^{3n}$ and reflectance $\vec{a}_\alb \in \mathbb{R}^{3n}$.
The shape $E_\id \in \mathbb{R}^{3n \times 80}$, reflectance $E_\alb \in \mathbb{R}^{3n \times 80}$, and expression $E_\expr \in \mathbb{R}^{3n \times 76}$ basis and the corresponding standard deviations $\sigma_\id \in \mathbb{R}^{80}$, $\sigma_\alb \in \mathbb{R}^{80}$, and $\sigma_\expr \in \mathbb{R}^{76}$ are given.
The model has $53$K vertices and $106$K faces.
A synthesized image $C_\mathcal{S}$ is generated through rasterization of the model under a rigid model transformation $\Phi \!  \left( \vec{v} \right)$ and the full perspective transformation $\Pi \!  \left( \vec{v} \right)$.
Illumination is approximated by the first three bands of Spherical Harmonics (SH) \cite{Ramamoorthi2001} basis functions, assuming Labertian surfaces and smooth distant illumination, neglecting self-shadowing.

Synthesis is dependent on the face model parameters $\vec{\alpha}$, $\vec{\beta}$, $\vec{\delta}$, the illumination parameters $\vec{\gamma}$, the rigid transformation $\mathbf{R}$, $\mathbf{t}$, and the camera parameters $\vec{\kappa}$ defining $\Pi$.
The vector of unknowns $\vec{\mathcal{P}}$ is the union of these parameters.

\section{Energy Formulation} \label{sec::energy}

Given a monocular input sequence, we reconstruct all unknown parameters $\mathcal{P}$ jointly with a robust variational optimization.
The proposed objective is highly non-linear in the unknowns and has the following components:
\begin{equation} \label{Eq:energy}
	E(\vec{\mathcal{P}}) \! = \! \underbrace{w_{col} E_{col}(\vec{\mathcal{P}}) + w_{lan} E_{lan}(\vec{\mathcal{P}})}_{data} + \underbrace{w_{reg} E_{reg}(\vec{\mathcal{P}})}_{prior} \enspace{.}
\end{equation}
The data term measures the similarity between the synthesized imagery and the input data in terms of photo-consistency $E_{col}$ and facial feature alignment $E_{lan}$.
The likelihood of a given parameter vector $\vec{\mathcal{P}}$ is taken into account by the statistical regularizer $E_{reg}$.
The weights $w_{col}$, $w_{lan}$, and $w_{reg}$ balance the three different sub-objectives. 
In all of our experiments, we set $w_{col}=1$, $w_{lan}=10$, and $w_{reg}=2.5\cdot10^{-5}$.
In the following, we introduce the different sub-objectives.

\paragraph*{Photo-Consistency}

In order to quantify how well the input data is explained by a synthesized image, we measure the photo-metric alignment error on pixel level:
\begin{equation} \label{Eq:color_term}
E_\col(\vec{\mathcal{P}}) = \frac{1}{|\mathcal{V}|}\sum_{\vec{p} \in \mathcal{V}}{\left\| C_\mathcal{S}(\vec{p}) - C_\mathcal{I}(\vec{p}) \right\|_2} \enspace{,}
\end{equation}
where $C_\mathcal{S}$ is the synthesized image, $C_\mathcal{I}$ is the input RGB image, and  $\vec{p} \! \in \! \mathcal{V}$ denote all visible pixel positions in $C_\mathcal{S}$.
We use the $\ell_{2,1}$-norm \cite{DingZHZ06} instead of a least-squares formulation to be robust against outliers.
In our scenario, distance in color space is based on $\ell_{2}$, while in the summation over all pixels an $\ell_{1}$-norm is used to enforce sparsity.

\paragraph*{Feature Alignment}

In addition, we enforce feature similarity between a set of salient facial feature point pairs detected in the RGB stream:
\begin{equation} \label{Eq:feature_term}
E_\lan(\vec{\mathcal{P}}) = \frac{1}{|\mathcal{F}|}\sum_{\vec{f}_j \in \mathcal{F}} {w_{\mathrm{conf}, j} \left\| \vec{f}_j - \Pi( \Phi \! \left( \vec{v}_j \right) \right\|^2_2} \enspace{.}
\end{equation}
To this end, we employ a state-of-the-art facial landmark tracking algorithm by \cite{Saragih2011}.
Each feature point $\vec{f}_j \in \mathcal{F} \subset\mathbb{R}^2$ comes with a detection confidence $w_{\mathrm{conf}, j}$ and corresponds to a unique vertex $\vec{v}_j = \mathcal{M}_{geo}(\boldsymbol \alpha, \boldsymbol \delta) \in \mathbb{R}^3$ of our face prior.
This helps avoiding local minima in the highly-complex energy landscape of $E_\col(\vec{\mathcal{P}})$.

\paragraph*{Statistical Regularization}

We enforce plausibility of the synthesized faces based on the assumption of a normal distributed population.
To this end, we enforce the parameters to stay statistically close to the mean:
\begin{equation} \scriptsize \label{Eq:regularization_term}
E_\reg(\vec{\mathcal{P}}) = 	\sum_{i=1}^{80} \left[	{\left( \frac{\vec{\alpha}_i}{\sigma_{\id, i}} \right)^2} + 
{\left( \frac{\vec{\beta}_i} {\sigma_{\alb, i}} \right)^2} \right] + 
\sum_{i=1}^{76} {\left( \frac{\vec{\delta}_i}{\sigma_{\expr, i}} \right)^2}  \enspace{.}
\end{equation}
This commonly-used regularization strategy prevents degenerations of the facial geometry and reflectance, and guides the optimization strategy out of local minima \cite{Blanz1999}.

\section{Data-parallel Optimization Strategy}
\label{sec::optimization}

The proposed robust tracking objective is a general unconstrained non-linear optimization problem.
We minimize this objective in real-time using a novel data-parallel GPU-based \emph{Iteratively Reweighted Least Squares} (IRLS) solver.
The key idea of IRLS is to transform the problem, in each iteration, to a non-linear least-squares problem by splitting the norm in two components:
$$||r(\vec{\mathcal{P}})||_2 = \underbrace{(||r(\vec{\mathcal{P}}_{old})||_2)^{-1}}_{constant}~ \cdot~ ||r(\vec{\mathcal{P}})||_2^2~.$$
Here, $r( \cdot )$ is a general residual and $\vec{\mathcal{P}}_{old}$ is the solution computed in the last iteration.
Thus, the first part is kept constant during one iteration and updated afterwards.
Close in spirit to \cite{thies2015realtime}, each single iteration step is implemented using the Gauss-Newton approach.
We take a single GN step in every IRLS iteration and solve the corresponding system of normal equations $\mathbf{J}^T\mathbf{J}\boldsymbol \delta^* = -\mathbf{J}^T \mathbf{F}$ based on PCG to obtain an optimal linear parameter update $\boldsymbol \delta^*$.
The Jacobian $\mathbf{J}$ and the systems' right hand side $-\mathbf{J}^T \mathbf{F}$ are precomputed and stored in device memory for later processing as proposed by Thies et al.~\cite{thies2015realtime}.
As suggested by \cite{Zollhoefer2014,thies2015realtime}, we split up the multiplication of the old descent direction $\vec{d}$ with the system matrix $\mathbf{J}^T\mathbf{J}$ in the PCG solver into two successive matrix-vector products.
Additional details regarding the optimization framework are provided in the supplemental material.

\section{Non-Rigid Model-Based Bundling}
\label{sec::bundling}

To estimate the identity of the actors in the heavily underconstrained scenario of monocular reconstruction, we introduce a non-rigid model-based bundling approach.
Based on the proposed objective, we jointly estimate all parameters over $k$ key-frames of the input video sequence.
The estimated unknowns are the global identity $\{\vec{\alpha}$, $\vec{\beta}\}$ and intrinsics $\vec{\kappa}$ as well as the unknown per-frame pose $\{\vec{\delta}^k$, $\mathbf{R}^k$, $\mathbf{t}^k\}_k$ and illumination parameters $\{\vec{\gamma}^k\}_k$.
We use a similar data-parallel optimization strategy as proposed for model-to-frame tracking, but jointly solve the normal equations for the entire keyframe set. 
For our non-rigid model-based bundling problem, the non-zero structure of the corresponding Jacobian is block dense.
Our PCG solver exploits the non-zero structure for increased performance (see additional document). 
Since all keyframes observe the same face identity under potentially varying illumination, expression, and viewing angle, we can robustly separate identity from all other problem dimensions. Note that we also solve for the intrinsic camera parameters of $\Pi$, thus being able to process uncalibrated video footage.
\section{Expression Transfer}
\label{sec::transfer}

To transfer the expression changes from the source to the target actor %in a person-specific manner, 
while preserving person-specificness in each actor's expressions, we propose a sub-space deformation transfer technique.
We are inspired by the deformation transfer energy of Sumner et al. \cite{Sumner2004}, but operate directly in the space spanned by the expression blendshapes.
This not only allows for the precomputation of the pseudo-inverse of the system matrix, but also drastically reduces the dimensionality of the optimization problem allowing for fast real-time transfer rates. 
Assuming source identity $\vec{\alpha}^S$ and target identity $\vec{\alpha}^T$ fixed, transfer takes as input the neutral $\vec{\delta}^S_N$, deformed source $\vec{\delta}^S$, and the neutral target $\vec{\delta}^T_N$ expression.
Output is the transferred facial expression  $\vec{\delta}^T$ directly in the reduced sub-space of the parametric prior.

As proposed by \cite{Sumner2004}, we first compute the source deformation gradients $\mathbf{A}_i \in \mathbb{R}^{3\times3}$ that transform the source triangles from neutral to deformed.
The deformed target $\hat{\vec{v}}_i = \vec{M}_{i}(\vec{\alpha}^T, \vec{\delta}^T)$ is then found based on the undeformed state $\vec{v}_i = \vec{M}_{i}(\vec{\alpha}^T, \vec{\delta}^T_N)$ by solving a linear least-squares problem.
Let $(i_0, i_1, i_2)$ be the vertex indices of the $i$-th triangle, $\mathbf{V} = \left[ \vec{v}_{i_1} - \vec{v}_{i_0}, \vec{v}_{i_2} - \vec{v}_{i_0} \right]$ and $\hat{\mathbf{V}} = \left[ \hat{\vec{v}}_{i_1} - \hat{\vec{v}}_{i_0}, \hat{\vec{v}}_{i_2} - \hat{\vec{v}}_{i_0} \right]$, then the optimal unknown target deformation $\vec{\delta}^T$ is the minimizer of:
\begin{equation} \footnotesize
E(\vec{\delta}^T) = \sum_{i=1}^{|\vec{F}|} \left|\left| 
\mathbf{A}_i \mathbf{V}
-\hat{\mathbf{V}}
\right|\right|^2_F~.
\end{equation}
This problem can be rewritten in the canonical least-squares form by substitution:
\begin{equation}
E(\vec{\delta}^T) = \left|\left| 
\mathbf{A} \vec{\delta}^T - \vec{b}
\right|\right|^2_2~.
\end{equation}
The matrix $\mathbf{A} \in \mathbb{R}^{6|\vec{F}| \times 76}$ is constant and contains the edge information of the template mesh projected to the expression sub-space.
Edge information of the target in neutral expression is included in the right-hand side $\vec{b} \in \mathbb{R}^{6|\vec{F}|}$.
$\vec{b}$ varies with $\vec{\delta}^S$ and is computed on the GPU for each new input frame.
The minimizer of the quadratic energy can be computed by solving the corresponding normal equations.
Since the system matrix is constant, we can precompute its \textit{Pseudo Inverse} using a Singular Value Decomposition (SVD).
Later, the small $76\times 76$ linear system is solved in real-time.
No additional smoothness term as in \cite{Sumner2004,Bouaziz2013} is needed, since the blendshape model implicitly restricts the result to plausible shapes and guarantees smoothness.

\section{Mouth Retrieval}
\label{sec::mouth}

For a given transferred facial expression, we need to synthesize a realistic target mouth region.
To this end, we retrieve and warp the best matching mouth image from the target actor sequence.
We assume that sufficient mouth variation is available in the target video.
It is also important to note that we maintain the appearance of the target mouth.
This leads to much more realistic results than either copying the source mouth region \cite{Vlasic2005,Dale2011} or using a generic $3$D teeth proxy \cite{Garrido2015,thies2015realtime}.

\begin{figure}
	\centering
	\includegraphics[width=\linewidth]{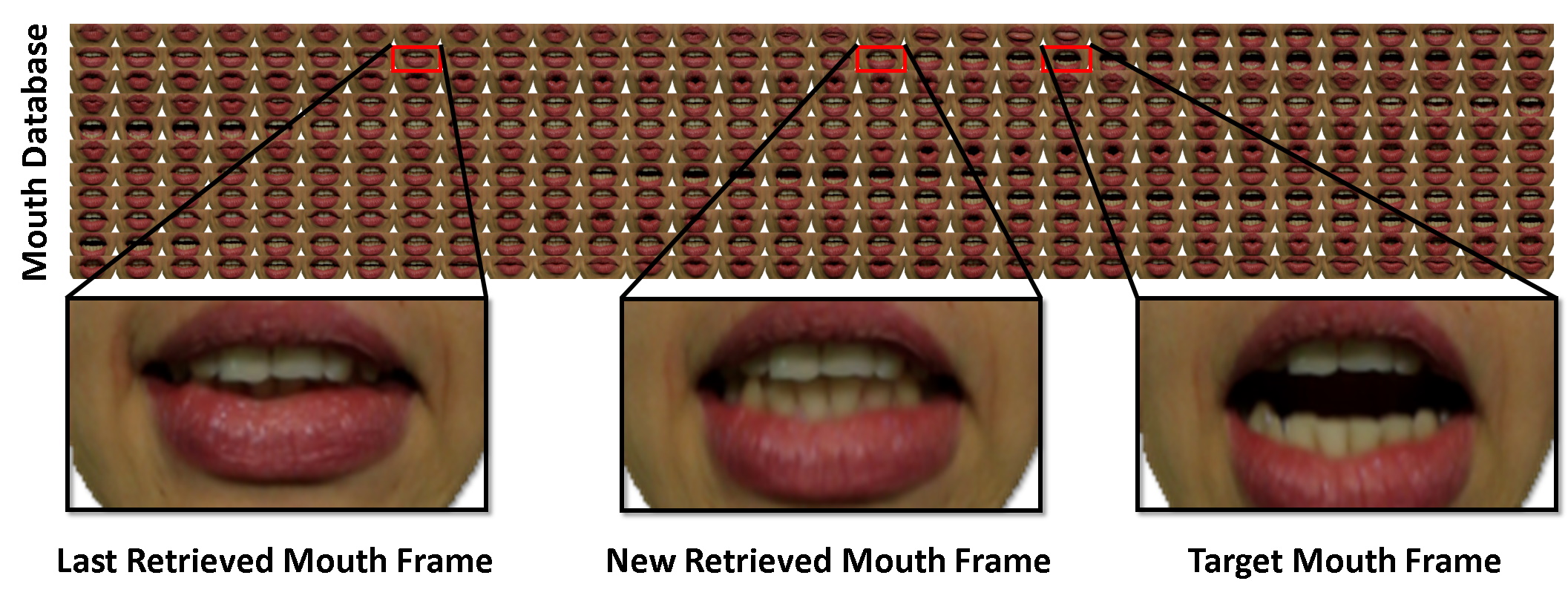}
	\caption{Mouth Retrieval: 
			we use an appearance graph to retrieve new mouth frames. In order to select a frame, we enforce similarity to the previously-retrieved frame while minimizing the distance to the target expression.
			%Based on an appearance graph the new retrieved mouth frame is chosen that is near to the old retrieved frame and the frame that is closest to the target expression.
			}
	\label{fig:mouthDatabase}
\end{figure}

Our approach first finds the best fitting target mouth frame based on a frame-to-cluster matching strategy with a novel feature similarity metric.
To enforce temporal coherence, we use a dense appearance graph to find a compromise between the last retrieved mouth frame and the target mouth frame (cf. Fig.~\ref{fig:mouthDatabase}).
We detail all steps in the following.

\paragraph{Similarity Metric}

Our similarity metric is based on geometric and photometric features.
The used descriptor $\vec{\mathcal{K}}=\{\mathbf{R}, \vec{\delta}, \mathcal{F}, \mathcal{L}\}$  of a frame is composed of the rotation $\mathbf{R}$, expression parameters $\vec{\delta}$, landmarks $\mathcal{F}$, and a Local Binary Pattern (LBP) $\mathcal{L}$.
We compute these descriptors $\vec{\mathcal{K}}^S$ for every frame in the training sequence.
The target descriptor $\vec{\mathcal{K}}^T$ consists of the result of the expression transfer and the LBP of the frame of the driving actor.
We measure the distance between a source and a target descriptor as follows:
{
\small
\begin{equation*}
D(\vec{\mathcal{K}}^T, \vec{\mathcal{K}}^S_t, t) = D_{p}(\vec{\mathcal{K}}^T, \vec{\mathcal{K}}^S_t) + D_{m}(\vec{\mathcal{K}}^T, \vec{\mathcal{K}}^S_t) + D_{a}(\vec{\mathcal{K}}^T, \vec{\mathcal{K}}^S_t, t)~.
\end{equation*}
}
The first term $D_{p}$ measures the distance in parameter space:
\begin{equation*}
D_{p}(\vec{\mathcal{K}}^T, \vec{\mathcal{K}}^S_t) = \|\vec{\delta}^T - \vec{\delta}^S_t \|_2^2 + \| \mathbf{R}^T - \mathbf{R}^S_t \|_F^2~.
\end{equation*}
The second term $D_{m}$ measures the differential compatibility of the sparse facial landmarks:
\begin{equation*}
D_m(\vec{\mathcal{K}}^T, \vec{\mathcal{K}}^S_t)  = \sum_{(i,j)\in \Omega}{ \left(\|\mathcal{F}^T_i - \mathcal{F}^T_j\|_2 - \|\mathcal{F}^S_{t,i} - \mathcal{F}^S_{t,j}\|_2 \right)^2}~.
\end{equation*}
Here, $\Omega$ is a set of predefined landmark pairs, defining distances such as between the upper and lower lip or between the left and right corner of the mouth.
The last term $D_{a}$ is an appearance measurement term composed of two parts:
\begin{equation*}
D_{a}(\vec{\mathcal{K}}^T, \vec{\mathcal{K}}^S_t, t) = D_{l}(\vec{\mathcal{K}}^T, \vec{\mathcal{K}}^S_t) + w_{c}(\vec{\mathcal{K}}^T, \vec{\mathcal{K}}^S_t) D_{c}(\tau, t)~.
\end{equation*}
$\tau$ is the last retrieved frame index used for the reenactment in the previous frame.
$D_{l}(\vec{\mathcal{K}}^T, \vec{\mathcal{K}}^S_t)$ measures the similarity based on LBPs that are compared via a \textit{Chi Squared Distance} (for details see \cite{Garrido2014}).
$D_{c}(\tau, t)$ measures the similarity between the last retrieved frame $\tau$ and the video frame $t$ based on RGB cross-correlation of the normalized mouth frames.
Note that the mouth frames are normalized based on the models texture parameterization (cf. Fig.~\ref{fig:mouthDatabase}).
To facilitate fast frame jumps for expression changes, we incorporate the weight
$w_{c}(\vec{\mathcal{K}}^T, \vec{\mathcal{K}}^S_t) = e^{-(D_m(\vec{\mathcal{K}}^T, \vec{\mathcal{K}}^S_t))^2}$.
We apply this frame-to-frame distance measure in a frame-to-cluster matching strategy, which enables real-time rates and mitigates high-frequency jumps between mouth frames.

\paragraph{Frame-to-Cluster Matching}
Utilizing the proposed similarity metric, we cluster the target actor sequence into $k=10$ clusters using a modified k-means algorithm that is based on the pairwise distance function $D$.
For every cluster, we select the frame with the minimal distance to all other frames within that cluster as a representative.
During runtime, we measure the distances between the target descriptor $\mathcal{K}^T$ and the descriptors of cluster representatives, and choose the cluster whose representative frame has the minimal distance as the new target frame.

\paragraph{Appearance Graph}
We improve temporal coherence by building a fully-connected appearance graph of all video frames.
The edge weights are based on the RGB cross-correlation between the normalized mouth frames, the distance in parameter space $D_p$, and the distance of the landmarks $D_m$.
The graph enables us to find an inbetween frame that is both similar to the last retrieved frame and the retrieved target frame (see Fig. \ref{fig:mouthDatabase}).
We compute this perfect match by finding the frame of the training sequence that minimizes the sum of the edge weights to the last retrieved and current target frame.
We blend between the previously-retrieved frame and the newly-retrieved frame in texture space on a pixel level after optic flow alignment.
Before blending, we apply an illumination correction that considers the estimated Spherical Harmonic illumination parameters of the retrieved frames and the current video frame.
Finally, we composite the new output frame by alpha blending between the original video frame, the illumination-corrected, projected mouth frame, and the rendered face model.

\section{Results}\label{sec::results}

\begin{figure*}
	\centering
	\ifdefined\LOWRES
	\includegraphics[width=\linewidth]{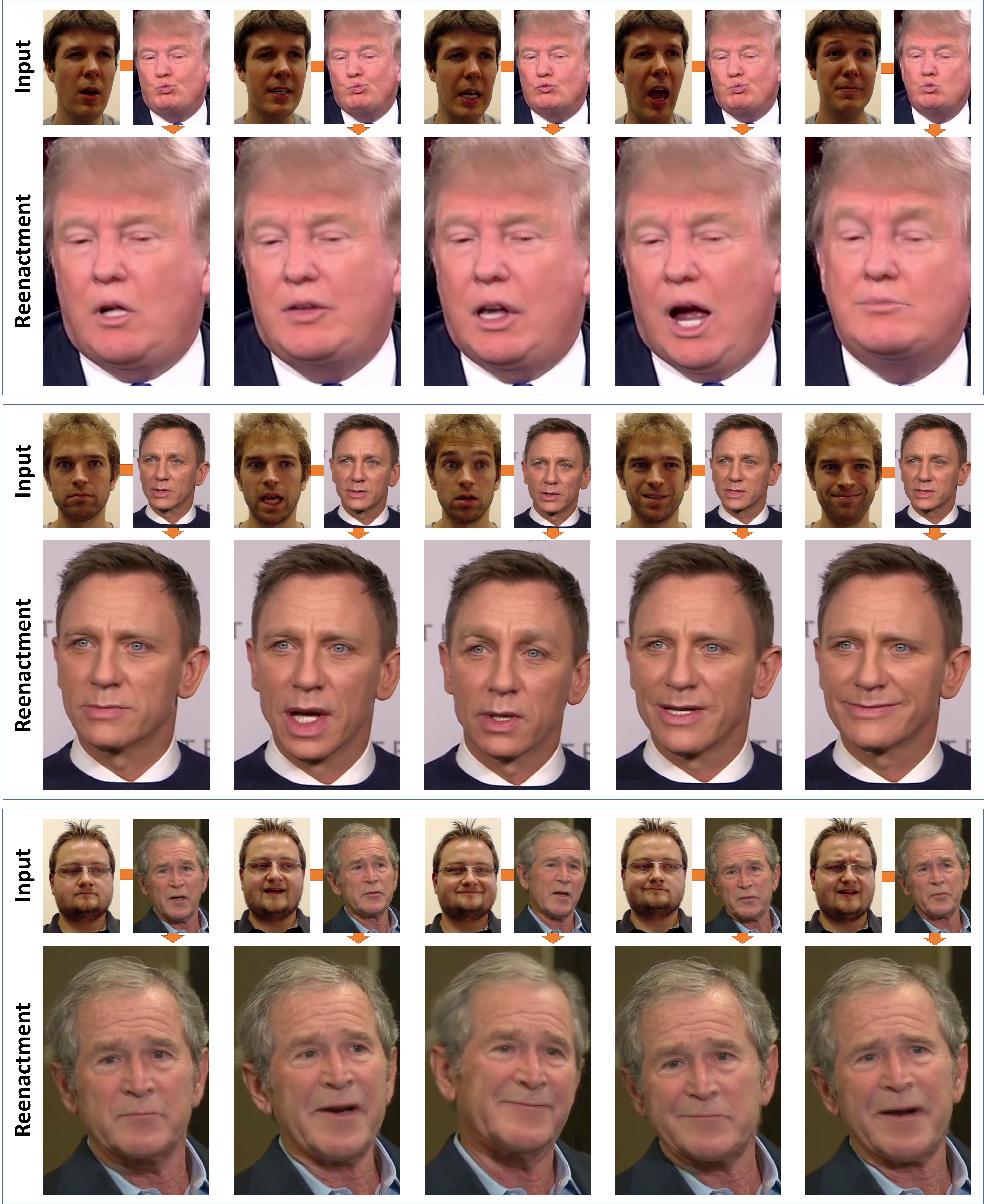}
	\else
	\includegraphics[width=\textwidth]{images/reenactmentResults}
	\fi
	\caption{Results of our reenactment system. Corresponding run times are listed in Table \ref{tab:timings}. The length of the source and resulting output sequences is 965, 1436, and 1791 frames, respectively; the length of the input target sequences is 431, 286, and 392 frames, respectively.}
	\label{fig:retargetingResults}
\end{figure*}

\paragraph{Live Reenactment Setup}
Our live reenactment setup consists of standard consumer-level hardware.
We capture a live video with a commodity webcam (source), and download monocular video clips from Youtube (target). 
In our experiments, we use a \textit{Logitech HD Pro C920} camera running at $30$Hz in a resolution of $640\times 480$; although our approach is applicable to any consumer RGB camera.
Overall, we show highly-realistic reenactment examples of our algorithm on a variety of target Youtube videos at a resolution of $1280 \times 720$. 
The videos show different subjects in different scenes filmed from varying camera angles;
each video is reenacted by several volunteers as source actors.
Reenactment results are generated at a resolution of $1280 \times 720$.
We show real-time reenactment results in Fig.~\ref{fig:retargetingResults} and in the accompanying video.

\paragraph{Runtime}
For all experiments, we use three hierarchy levels %on the input data 
for tracking (source and target).
In pose optimization, we only consider the second and third level, where we run one and seven Gauss-Newton steps, respectively.
Within a Gauss-Newton step, we always run four PCG steps.
In addition to tracking, our reenactment pipeline has additional stages whose timings are listed in Table \ref{tab:timings}.
Our method runs in real-time on a commodity desktop computer with an NVIDIA Titan X and an Intel Core i7-4770.

\begin{table}
	{
		\footnotesize
		\centering
		\begin{tabular}{|c|c|c|c|c|c|}
			\hline
			 \multicolumn{2}{|c|}{CPU} & \multicolumn{3}{c|}{GPU} & FPS \\
			\hline
			 SparseFT & MouthRT & DenseFT & DeformTF & Synth &  \\
			\hline
			 5.97ms & 1.90ms & 22.06ms & 3.98ms & 10.19ms & \textbf{27.6Hz} \\ %daniel-trump
			\hline
			 4.85ms & 1.50ms & 21.27ms & 4.01ms & 10.31ms & \textbf{28.1Hz} \\ %maddi-craig
			\hline
			 5.57ms & 1.78ms & 20.97ms & 3.95ms & 10.32ms & \textbf{28.4Hz} \\ %justus-bush
			 \hline
		\end{tabular}
	}
	\caption{
		Avg. run times for the three sequences of Fig.~ \protect{\ref{fig:retargetingResults}}, from top to bottom. Standard deviations w.r.t. the final frame rate are $0.51$, $0.56$, and $0.59$ fps, respectively. Note that CPU and GPU stages run in parallel.}
	\label{tab:timings}
\end{table}

\paragraph{Tracking Comparison to Previous Work} 
Face tracking alone is not the main focus of our work, but the following comparisons show that our tracking is on par with or exceeds the state of the art.

\begin{figure}
	\centering
	\ifdefined\LOWRES
	\includegraphics[width=\linewidth]{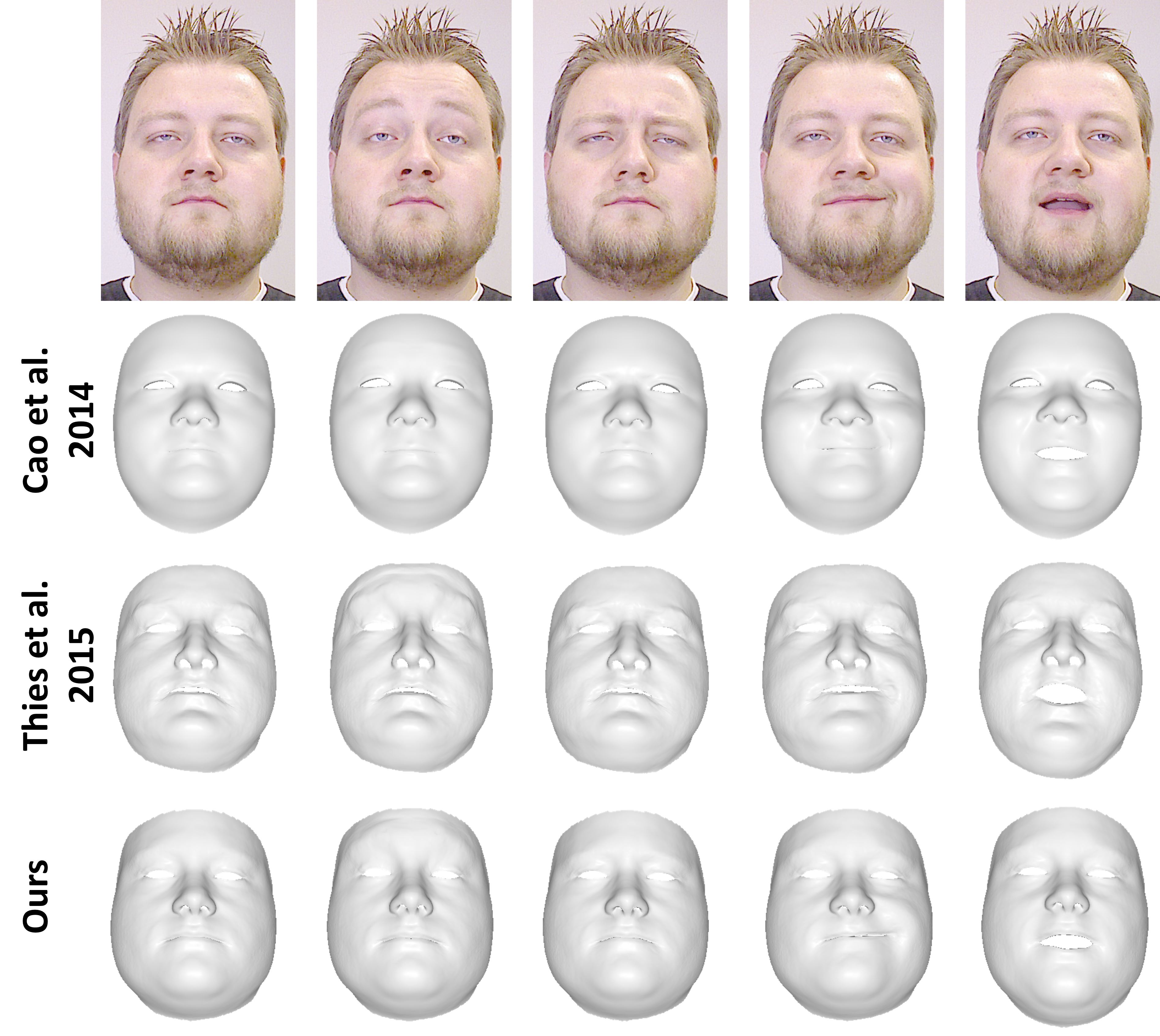}
	\else
	\includegraphics[width=\textwidth]{images/cao_comparison}
	\fi
	\caption{Comparison of our RGB tracking to Cao et al.~\cite{Cao2014}, and to RGB-D tracking by Thies et al.~\cite{thies2015realtime}.}
	\label{fig:caoTracking}
\end{figure}

\begin{figure}
	\centering
	\ifdefined\LOWRES
	\includegraphics[width=\linewidth]{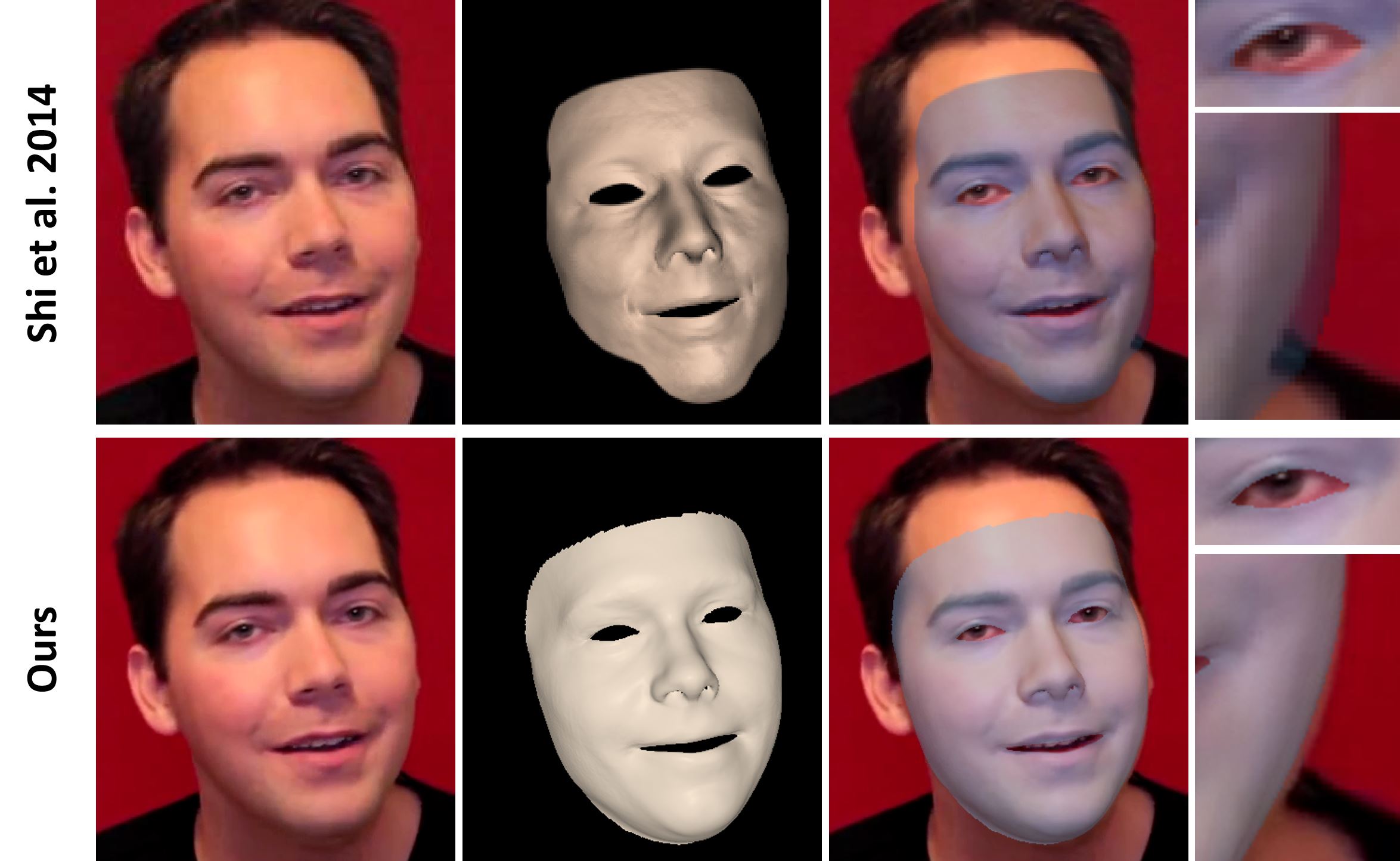}
	\else
	\includegraphics[width=\textwidth]{images/shi_comparison}
	\fi
	\caption{Comparison of our tracking to Shi et al. \cite{Shi2014}. From left to right: RGB input, reconstructed model, overlay with input, close-ups on eye and cheek. Note that Shi et al. perform shape-from-shading in a post process.}
	\label{fig:shiTracking}
\end{figure}

\textit{Shi et al.~2014 \cite{Shi2014}:}
They capture face performances offline from monocular unconstrained RGB video.
The close-ups in Fig.~\ref{fig:shiTracking} show that our online approach yields a closer face fit, particularly visible at the silhouette of the input face.
We believe that our new dense non-rigid bundle adjustment leads to a better shape identity estimate than their sparse approach.

\textit{Cao et al.~2014 \cite{Cao2014}:}
They capture face performance from monocular RGB in real-time. 
In most cases, our and their method produce similar high-quality results (see Fig.~\ref{fig:caoTracking}); our identity and expression estimates are slightly more accurate though.

\textit{Thies et al.~2015 \cite{thies2015realtime}:}
Their approach captures face performance in real-time from RGB-D, Fig. \ref{fig:caoTracking}.
Results of both approaches are similarly accurate; but our approach does not require depth data.

\begin{figure}
	\centering
	\ifdefined\LOWRES
	\includegraphics[width=0.95\linewidth]{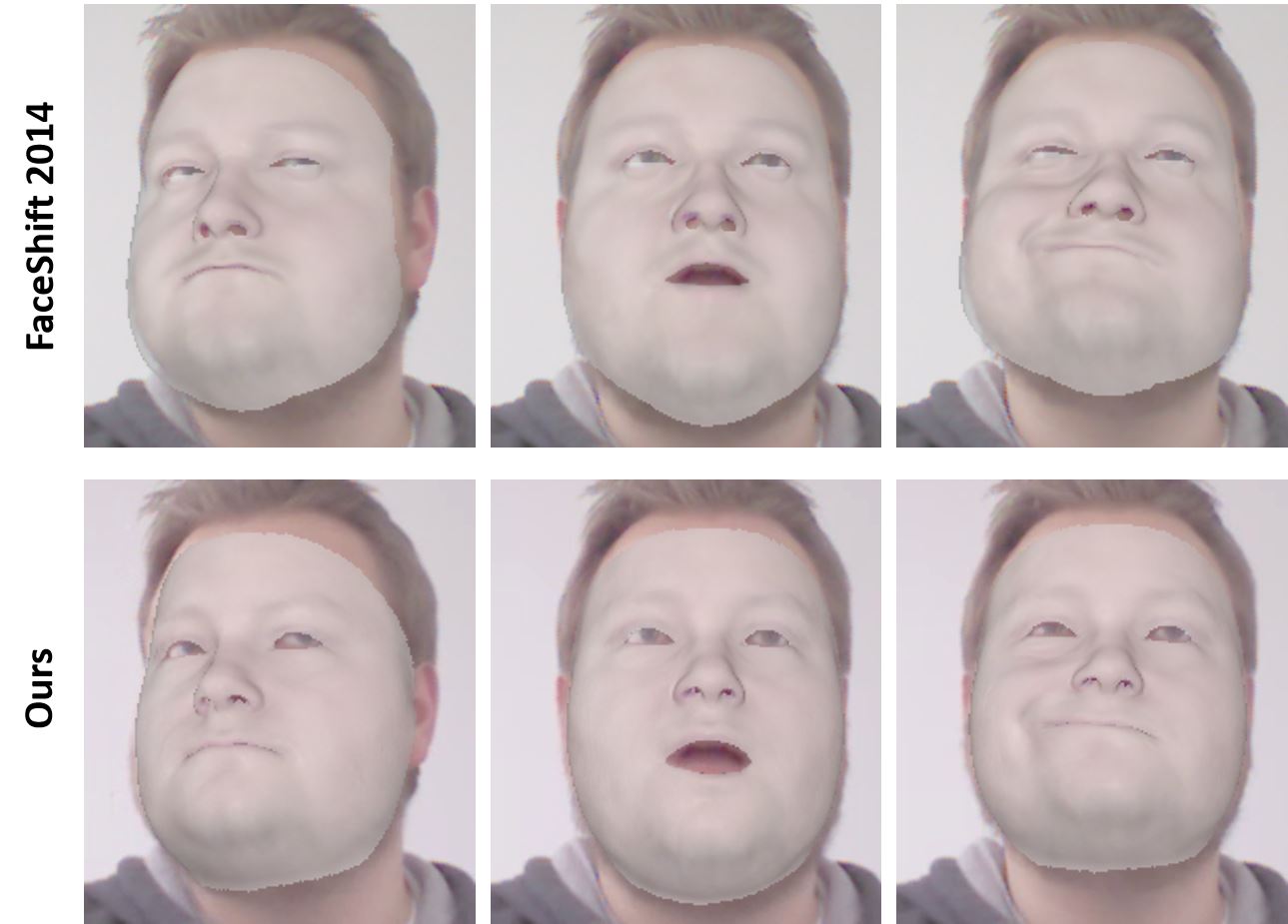}
	\else
	\includegraphics[width=0.95\textwidth]{images/faceshift_comparison}
	\fi
	\caption{Comparison against \textit{FaceShift} RGB-D tracking.}
	\label{fig:faceshift}
\end{figure}

\textit{FaceShift 2014:}
We compare our tracker to the commercial real-time RGB-D tracker from \textit{FaceShift}, which is based on the work of Weise et al. \cite{Weise2011}. 
Fig.~\ref{fig:faceshift} shows that we obtain similar results from RGB only.

\begin{figure}
	\centering
	\includegraphics[width=0.75\linewidth]{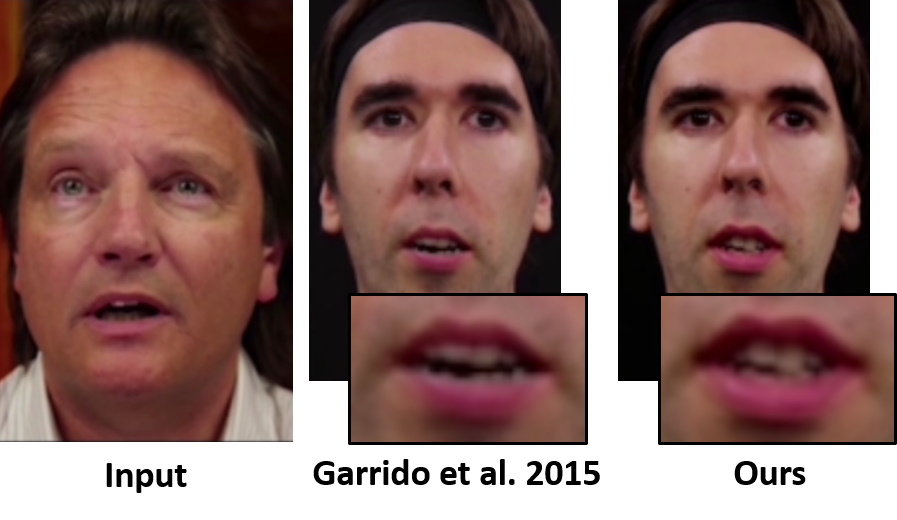}
	\vspace{-0.4cm}
	\caption{Dubbing: Comparison to Garrido et al.~\cite{Garrido2015}.}
	\label{fig:dubbing}
\end{figure}

\begin{figure}
	\centering
	\includegraphics[width=0.9\linewidth]{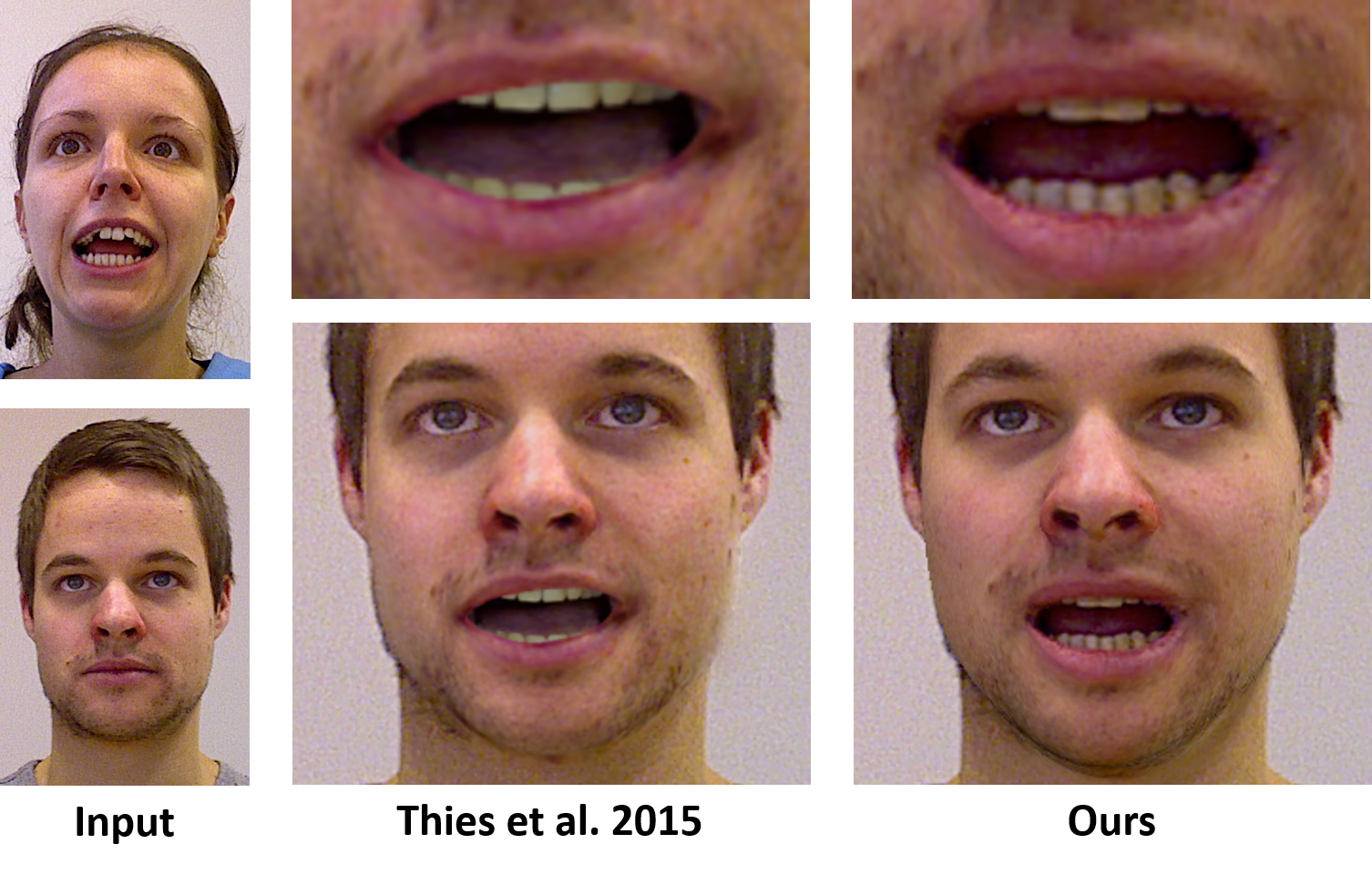}
	\vspace{-0.4cm}
	\caption{Comparison of the proposed RGB reenactment to the RGB-D reenactment of Thies et al. \cite{thies2015realtime}.}
	\label{fig:reenactmentOld}
\end{figure}

\paragraph{Reenactment Evaluation}
In Fig.~\ref{fig:dubbing}, we compare our approach against state-of-the art reenactment by Garrido et al. \cite{Garrido2015}. 
Both methods provide highly-realistic reenactment results; however, their method is fundamentally offline, as they require all frames of a sequence to be present at any time. %; their runtime is also orders of magnitude slower than ours.
In addition, they rely on a generic geometric teeth proxy which in some frames makes reenactment less convincing.
In Fig.~\ref{fig:reenactmentOld}, we compare against the work by Thies et al. \cite{thies2015realtime}.
Runtime and visual quality are similar for both approaches; however, their geometric teeth proxy leads to undesired appearance changes in the reenacted mouth.
Moreover, Thies et al. use an RGB-D camera, which limits the application range; they cannot reenact Youtube videos.
We show additional comparisons in the supplemental material against Dale et al.~\cite{Dale2011} and Garrido et al.~\cite{Garrido2014}.

\section{Limitations}\label{sec::limitation}

The assumption of Lambertian surfaces and smooth illumination is limiting, and may lead to artifacts in the presence of hard shadows or specular highlights; a limitation shared by most state-of-the-art methods. Scenes with face occlusions by long hair and a beard are challenging. 
Furthermore, we only reconstruct and track a low-dimensional blendshape model (76 expression coefficients), which omits fine-scale static and transient surface details.
Our retrieval-based mouth synthesis assumes sufficient visible expression variation in the target sequence.
On a too short sequence, or when the target remains static, we cannot learn the person-specific mouth behavior.
In this case, temporal aliasing can be observed, as the target space of the retrieved mouth samples is too sparse. %to allow for smooth frame-to-frame transitions.
Another limitation is caused by our hardware setup (webcam, USB, and PCI), which introduces a small delay of $\approx3$ frames.
Specialized hardware could resolve this, but our aim is a setup with commodity hardware.

\section{Conclusion}\label{sec::conclusion}

The presented approach is the first real-time facial reenactment system that requires just monocular RGB input.
Our live setup enables the animation of legacy video footage -- e.g., from Youtube -- in real time.
Overall, we believe our system will pave the way for many new and exciting applications in the fields of VR/AR, teleconferencing, or on-the-fly dubbing of videos with translated audio.

\section*{Acknowledgements}
We would like to thank Chen Cao and Kun Zhou for the blendshape models and comparison data, as well as Volker Blanz, Thomas Vetter, and Oleg Alexander for the provided face data. The facial landmark tracker was kindly provided by TrueVisionSolution.
We thank Angela Dai for the video voice over and Daniel Ritchie for video reenactment.
This research is funded by the German Research Foundation (DFG), grant GRK-1773 Heterogeneous Image Systems, the ERC Starting Grant 335545 CapReal, and the Max Planck Center for Visual Computing and Communications (MPC-VCC).
We also gratefully acknowledge the support from NVIDIA Corporation for hardware donations.

{\small
\bibliographystyle{ieee}
\bibliography{face}
}

\newpage
\begin{appendix}
	
\section{Optimization Framework}
\label{sec::optimization}

Our Gauss-Newton optimization framework is based on the work of Thies et al. \cite{thies2015realtime}.
Our aim is to include every visible pixel $\vec{p} \! \in \! \mathcal{V}$ in $C_\mathcal{S}$ in the optimization process. %; i.e., one residual term for each of them. (not true, 3 per pixel)
To this end, we gather all visible pixels in the synthesized image using a parallel prefix scan.
The computation of the Jacobian $J$ of the residual vector $F$ and the gradient $J^TF$ of the energy function are then parallelized across all GPU processors.
This parallelization is feasible since all partial derivatives and gradient entries with respect to a variable can be computed independently.
During evaluation of the gradient, all components of the Jacobian are computed and stored in global memory. 
In order to evaluate the gradient, we use a two-stage reduction to sum-up all local per pixel gradients.
Finally, we add the regularizer and the sparse feature term to the Jacobian and the gradient.
Using the computed Jacobian $J$ and the gradient $J^TF$, we solve the corresponding normal equation
$J^TJ \Delta x = -J^TF$ for the parameter update $\Delta x$ using a preconditioned conjugate gradient (PCG) method. We apply a Jacobi preconditioner that is precomputed during the evaluation of the gradient.
To avoid the high computational cost of $J^TJ$, our GPU-based PCG method splits up the computation of $J^TJp$ into two successive matrix-vector products.

In order to increase convergence speed and to avoid local minima, we use a coarse-to-fine hierarchical optimization strategy.
During online tracking, we only consider the second and third level, where we run one and seven Gauss-Newton steps on the respective level.
Within a Gauss-Newton step, we always run four PCG iterations.

Our complete framework is implemented using DirectX for rendering and DirectCompute for optimization.
The joint graphics and compute capability of DirectX11 enables the processing of rendered images by the graphics pipeline without resource mapping overhead.
In the case of an \textit{analysis-by-synthesis approach} like ours, this is essential to runtime performance, since many rendering-to-compute switches are required.

\section{Non-rigid Bundling}
\label{sec::bundling}

\begin{figure}[h]
	\centering
	\includegraphics[width=\linewidth]{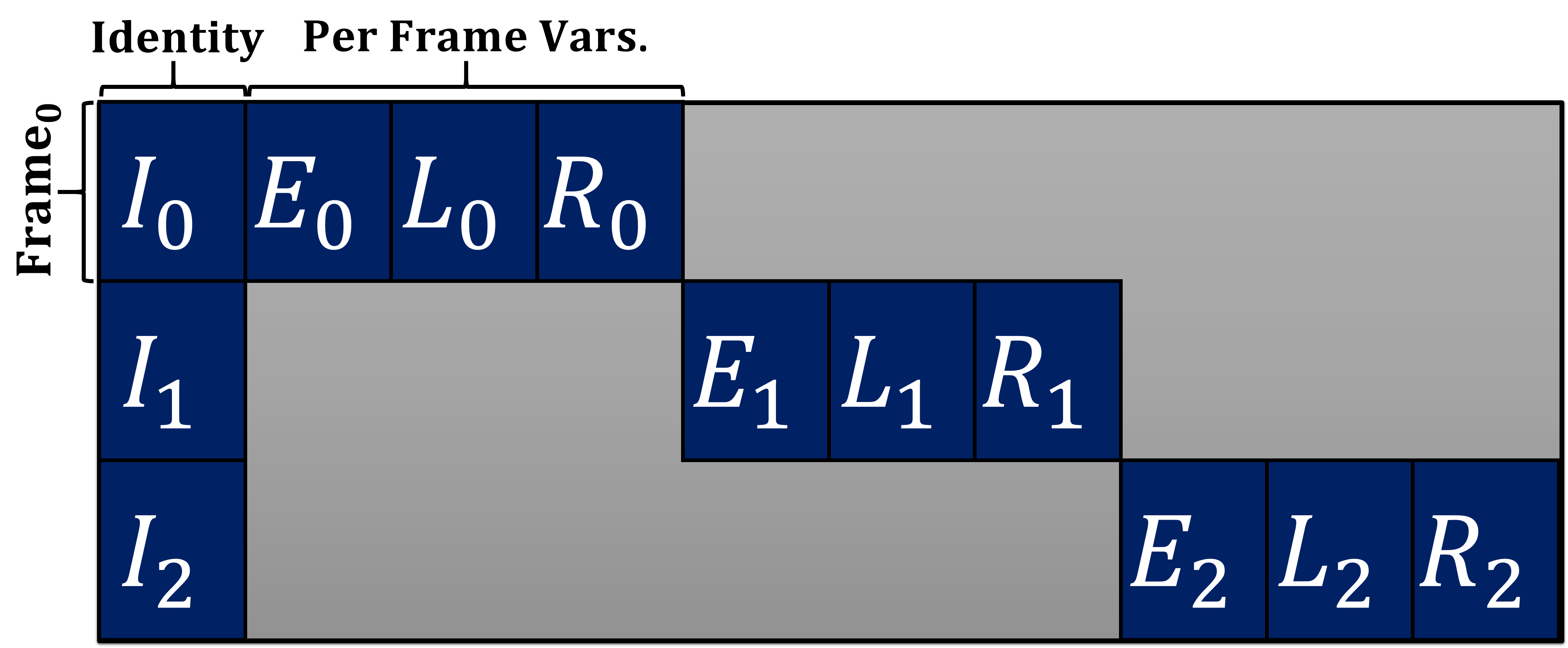}
	\caption{Non-zero structure of the Jacobian matrix of our non-rigid model-based bundling approach for three key-frames. Where $I_i, E_i, L_i, R_i$ are the $i$-th per frame Jacobian matrices of the identity, expression, illumination, and rigid pose parameters.}
	\label{fig:jacobian_bundling}
\end{figure}

For our non-rigid model-based bundling problem, the non-zero structure of the corresponding Jacobian is block dense.
We visualize its non-zero structure, which we exploit during optimization, in Fig.~\ref{fig:jacobian_bundling}.
In order to leverage the sparse structure of the Jacobian, we adopt the Gauss-Newton framework as follows:
we modify the computation of the gradient $J^T(\vec{\mathcal{P}}) \cdot F(\vec{\mathcal{P}})$ and the matrix vector product $J^T(\vec{\mathcal{P}}) \cdot J(\vec{\mathcal{P}}) \cdot \vec{x}$ that is used in the PCG method.
To this end, we define a promoter function $\Psi_f:\mathbb{R}^{|\mathcal{P}_{global}|+|\mathcal{P}_{local}|} \rightarrow \mathbb{R}^{|\mathcal{P}_{global}|+k \cdot |\mathcal{P}_{local}|}$ that lifts a per frame parameter vector to the parameter vector space of all frames ($\Psi^{-1}_f$ is the inverse of this promoter function).
$\mathcal{P}_{global}$ are the global parameters that are shared over all frames, such as the identity parameters of the face model and the camera parameters.
$\mathcal{P}_{local}$ are the local parameters that are only valid for one specific frame (i.e., facial expression, rigid pose and illumination parameters).
Using the promoter function $\Psi_f$ the gradient is given as
\begin{equation*}
J^T(\vec{\mathcal{P}}) \cdot F(\vec{\mathcal{P}}) = \sum_{f=1}^k {\Psi_f(J_f^T(\Psi^{-1}_f(\vec{\mathcal{P}})) \cdot F_f(\Psi^{-1}_f(\vec{\mathcal{P}})))},
\end{equation*}
where $J_f$ is the per-frame Jacobian matrix and $F_f$ the corresponding residual vector.

As for the parameter space, we introduce another promoter function $\hat{\Psi}_f$ that lifts a local residual vector to the global residual vector. 
In contrast to the parameter promoter function, this function varies in every Gauss-Newton iteration since the number of residuals might change.
As proposed in \cite{Zollhoefer2014,thies2015realtime}, we split up the computation of $J^T(\vec{\mathcal{P}}) \cdot J(\vec{\mathcal{P}}) \cdot \vec{x}$ into two successive matrix vector products, where the second multiplication is analogue to the computation of the gradient.
The first multiplication is as follows:
\begin{equation*}
J(\vec{\mathcal{P}}) \cdot \vec{x} = \sum_{f=1}^k \hat{\Psi}_f \left( J_f(\Psi^{-1}_f(\vec{\mathcal{P}})) \cdot \Psi^{-1}_f(\vec{x}) \right)
\end{equation*}
Using this scheme, we are able to efficiently solve the normal equations.

\begin{figure}[h]
	\centering
	\includegraphics[width=\linewidth]{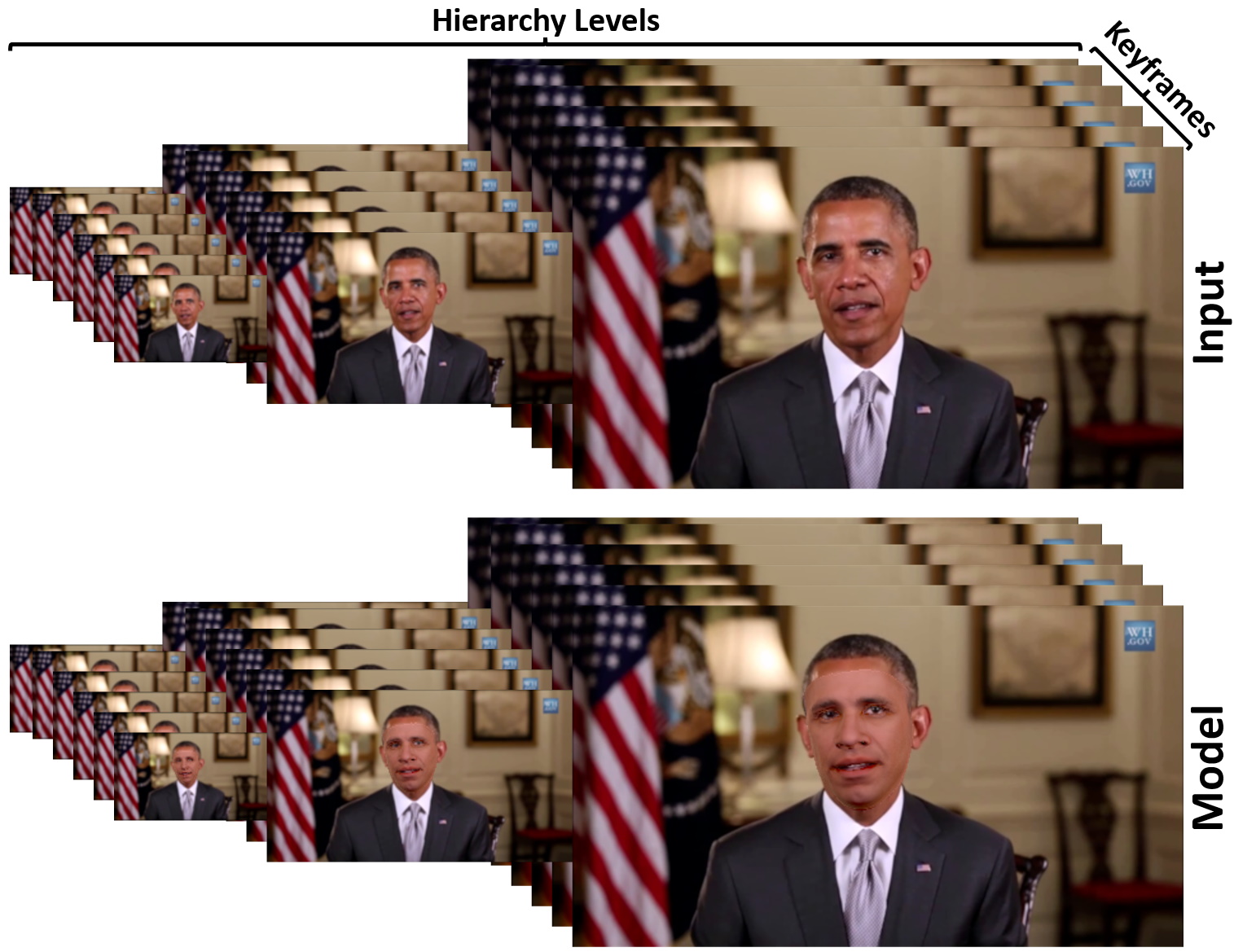}
	\caption{Non-rigid model-based bundling hierarchy: the top row shows the hierarchy of the input video and the second row the overlaid face model.}
	\label{fig:bundling}
\end{figure}

The Gauss-Newton framework is embedded in a hierarchical solution strategy (see Fig.~\ref{fig:bundling}).
This hierarchy allows to prevent convergence to local minima.
We start optimizing on a coarse level and propagate the solution to the next finer level using the parametric face model.
In our experiments we used three levels with $25$, $5$, and $1$ Gauss-Newton iterations for the coarsest, the medium and the finest level respectively, each with $4$ PCG steps.
Our implementation is not restricted to the number $k$ of used keyframes.
The processing time is linear in the number of keyframes.
In our experiments we used $k=6$ keyframes to estimate the identity parameters resulting in a processing time of a few seconds ($\sim20s$).

\section{Reenactment Evaluation}
\label{sec::reenactment}

\begin{figure}[h]
	\centering
	\includegraphics[width=0.8\linewidth]{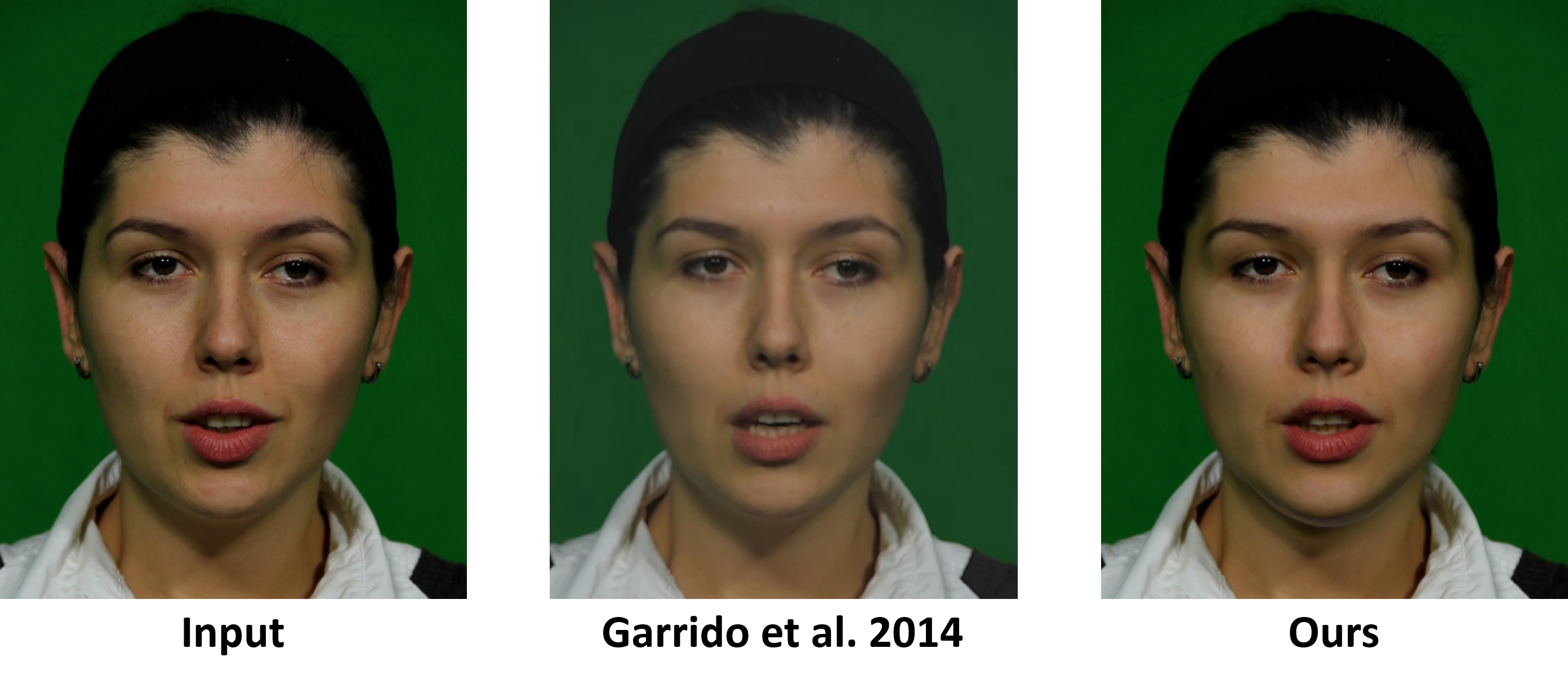}
	\caption{Self-Reenactment comparison to Garrido et al. \cite{Garrido2014}. The expression of the actress is transferred to a recorded video of herself.}
	\label{fig:reenactmentPablo}
\end{figure}

\begin{figure}[h]
	\centering
	\includegraphics[width=\linewidth]{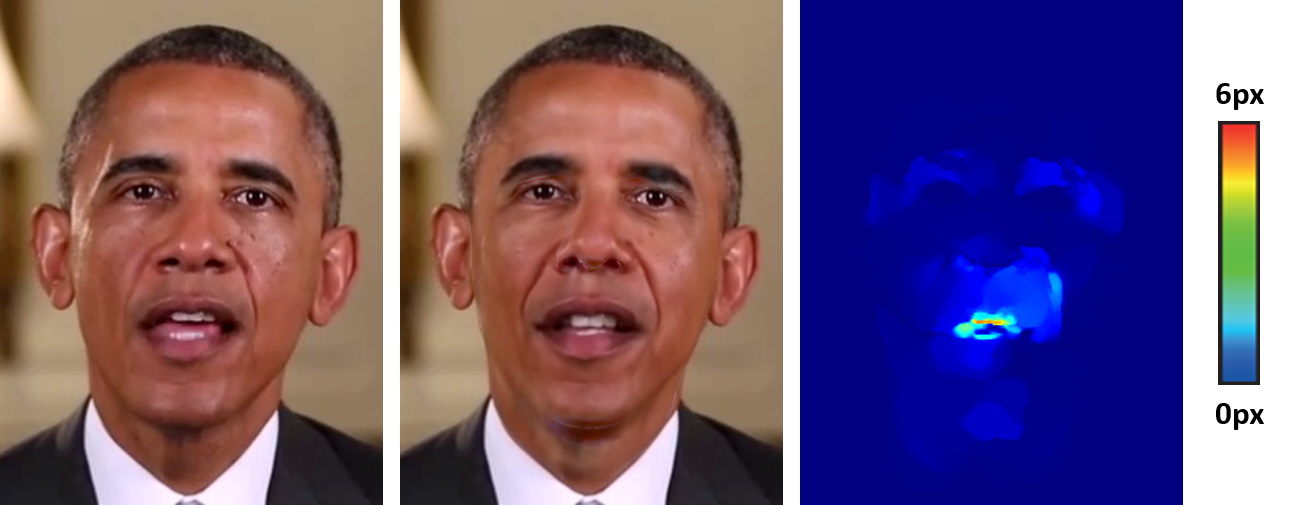}
	\caption{Self-Reenactment / Cross-Validation; from left to right: input frame (ground truth), resulting self-reenactment, and the photometric error.}
	\label{fig:selfReenactment}
\end{figure}

In addition to the results in the main paper \cite{thies2016face}, we compare our method to other existing reenactment pipelines.
Fig.~\ref{fig:reenactmentPablo} shows a self-reenactment scenario (i.e., the source and the target actor is the same person) in comparison to Garrido et al. \cite{Garrido2014}.
Our online approach is able to achieve similar or better quality as the offline approach of Garrido et al. \cite{Garrido2014}.
\begin{figure}[h]
	\centering
	\includegraphics[width=\linewidth]{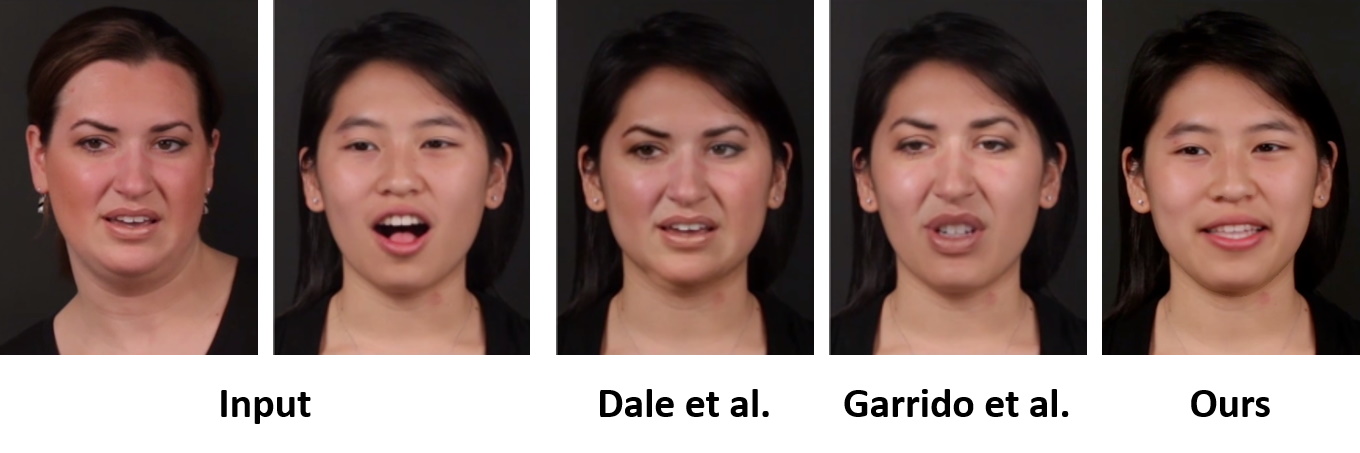}
	\caption{Comparison to Dale et al. \cite{Dale2011} and Garrido et al. \cite{Garrido2014}.
		The expression of the left input actor is transferred to the right input actor without changing the person's identity.}
	\label{fig:reenactmentDale}
\end{figure}
In Fig.~\ref{fig:reenactmentDale}, we show a comparisons to Dale et al. \cite{Dale2011} and Garrido et al. \cite{Garrido2014}.
Note that both methods do not preserve the identity of the target actor outside of the self-reenactment scenario.
In contrast, our method preserves the identity and alters the expression with respect to the source actor, which enables more plausible results.
We evaluate the presented reenactment method by measuring the photometric error between the input sequence and the self-reenactment of an actor using cross-validation (see Fig.~\ref{fig:selfReenactment}).
The first 1093 frames of the video are used to retrieve mouth interiors (training data).
Thus, self-reenactment of the first half results in a small mean photometric error of 0.33 pixels (0.157px std.Dev.) measured via optical flow.
In the second half (frames 1093-2186) of the video, the photometric error increases to a mean value of 0.42 pixels (0.17px std.Dev.).

\begin{table}[h]
	{
		\footnotesize
		\centering
		\begin{tabular}{|l|}
			\hline
			Obama - Celebrating Independence Day \\ \href{https://www.youtube.com/watch?v=d-VaUaTF3_k}{https://www.youtube.com/watch?v=d-VaUaTF3\_k} \\
			\hline
			Donald Trump - Interview: 'I Love China' - Morning Joe - MSNBC \\ \href{https://www.youtube.com/watch?v=Tsh_V3U7EfU}{https://www.youtube.com/watch?v=Tsh\_V3U7EfU} \\
			\hline
			Daniel Craig - Interview on the new James Bond Movie \\ \href{https://www.youtube.com/watch?v=8ZbCf7szjXg}{https://www.youtube.com/watch?v=8ZbCf7szjXg} \\
			\hline
			Putin - New Year's Address to the Nation \\ \href{https://www.youtube.com/watch?v=8_JxKKY7I_Y}{https://www.youtube.com/watch?v=8\_JxKKY7I\_Y} \\
			\hline
			Arnold Schwarzenegger - Terminator: Genisys \\ \href{https://www.youtube.com/watch?v=p6CJx_ZbaG4}{https://www.youtube.com/watch?v=p6CJx\_ZbaG4} \\
			\hline
			Vocal Coach Ken Taylor - How to Sing Well \\ \href{https://www.youtube.com/watch?v=KDYaACGUt3k}{https://www.youtube.com/watch?v=KDYaACGUt3k} \\
			\hline
		\end{tabular}
	}
	\caption{Youtube Video References.}
	\label{tab:links}
\end{table}

\newpage

\section{List of Mathematical Symbols}
\label{appendix::symbols}

\begin{table}[h]
	{
		\centering
\begin{tabular}{|c|l|}
	\hline
	\multicolumn{1}{|c|}{\textbf{Symbol}} & \multicolumn{1}{|c|}{\textbf{Description}}\\
	\hline
	\hline	
	
	% mouth retrieval
	$\vec{ \mathcal{K} }$ & feature descriptor \\ \hline
	$\mathcal{L}$ & Local Binary Pattern \\ \hline
	$t$ & timestep \\ \hline
	$D  \!  \left( \vec{ \mathcal{K} }^T, \vec{ \mathcal{K} }^S, t \right)$ & distance measure \\ \hline
	$D_p  \!  \left( \vec{ \mathcal{K} }^T, \vec{ \mathcal{K} }^S_t \right)$ & distance measure in parameter space \\ \hline
	$D_m  \!  \left( \vec{ \mathcal{K} }^T, \vec{ \mathcal{K} }^S_t \right)$ & distance measure of facial landmarks \\ \hline
	$D_a  \!  \left( \vec{ \mathcal{K} }^T, \vec{ \mathcal{K} }^S_t, t \right)$ & distance measure of appearance \\ \hline
	$D_l  \!  \left( \vec{ \mathcal{K} }^T, \vec{ \mathcal{K} }^S_t \right)$ & Chi Squared Distance of LBPs \\ \hline
	$D_c  \!  \left( \tau, t \right)$ & cross-correlation between frame $\tau$ and $t$ \\ \hline
	$\tau_k$ & $k$-th previous retrieved frame index \\ \hline
	$w_c  \!  \left( \vec{ \mathcal{K} }^T, \vec{ \mathcal{K} }^S_t \right)$ & frame weight \\ \hline

	\hline
	\hline	

	$\Phi  \!  \left( \vec{ \mathcal{P} } \right)$ & parameter promoter function \\ \hline
	$\hat{\Phi}  \!  \left( F \left( \vec{ \mathcal{P} } \right) \right)$ & residual promoter function \\ \hline
\end{tabular}
}
\end{table}

\newpage

\begin{table}[h]
	{
		\centering
\begin{tabular}{|c|l|}
	\hline
	\multicolumn{1}{|c|}{\textbf{Symbol}} & \multicolumn{1}{|c|}{\textbf{Description}}\\
	\hline
	\hline
	% face model
	$\vec{\alpha}, \vec{\beta}, \vec{\delta}$ & shape, albedo, expression parameters\\ 					\hline
	$\mathcal{M}_\geo, \mathcal{M}_\alb$ & parametric face model \\ \hline
	$\vec{a}_\id, \vec{a}_\alb$ & average shape, albedo \\ \hline
	$E_\id, E_\alb, E_\expr$ & shape, albedo, expression basis \\ \hline
	$\vec{\sigma}_\id, \vec{\sigma}_\alb, \vec{\sigma}_\expr$ & std. dev. shape, albedo, expression \\ \hline
	$\vec{F}$ & triangle set of the model \\  \hline
	$n$ & number of vertices\\ \hline
	$\vec{v}_j$ & vertex of the face model \\   \hline
	% illumination
	$\vec{\gamma}$ & illumination parameters \\ \hline	
	% transformations
	$\Phi \!  \left( \vec{v} \right)$ & model-to-world transformation \\ \hline
	$\vec{R}$ & rotation \\ \hline
	$\vec{t}$ & translation \\ \hline
	$\Pi \!  \left( \vec{v} \right)$ & full perspective projection \\ \hline
	$\kappa$ & camera parameters defining $\Pi \!  \left( \vec{v} \right)$ \\ \hline
	% parameters
	$\vec{\mathcal{P}}$ & vector of all parameters \\ \hline
	% input data	
	$C_\mathcal{I}$ & input color \\ \hline
	$C_\mathcal{S}$ & synth. color \\ \hline	
	$\mathcal{V}$ & set of valid pixels \\ \hline
	$\vec{p}$ & integer pixel location \\ \hline
	$\mathcal{F}$ & set of detected features \\ \hline
	$\vec{f}_j$ & $j$-th feature point \\ \hline
	$w_{\mathrm{conf}, j}$ & confidence of $j$-th feature point \\ \hline
	% energy function
	$E  \!  \left( \vec{\mathcal{P}} \right)$ & energy function \\ \hline
	$E_{col}  \!  \left( \vec{\mathcal{P}} \right)$ & photo-consistency term \\ \hline
	$E_{lan}  \!  \left( \vec{\mathcal{P}} \right)$ & feature alignment term \\ \hline
	$E_{reg}  \!  \left( \vec{\mathcal{P}} \right)$ & statistical regularization \\ \hline
	$w_{col}, w_{lan}, w_{reg}$ & energy term weights \\ \hline	
	% optimizer
	$r(\vec{\mathcal{P}})$ & a general residual vector \\ \hline
	$J(\vec{\mathcal{P}})$ & jacobian matrix \\ \hline
	$F(\vec{\mathcal{P}})$ & residual vector \\ \hline \hline

	% expression transfer
	$A_i$ & deformation gradient of triangle $i$ \\ \hline
	$\hat{\vec{v}}_i$ & deformed vertex \\ \hline
	$\vec{V}$ & triangle spanning vectors \\ \hline
	$\hat{\vec{V}}$ & deformed triangle spanning vectors \\ \hline
	$E  \!  \left( \vec{\delta}^T \right)$ & deformation transfer energy \\ \hline
	$A$ & system matrix of the transfer energy \\ \hline
	$b$ & rhs of the transfer energy \\ \hline
\end{tabular}
}
\end{table}

\end{appendix}

\end{document}